\documentclass[11pt]{article}

\usepackage[final]{acl}
\usepackage{times}
\usepackage{latexsym}
\usepackage[T1]{fontenc}
\usepackage[utf8]{inputenc}
\usepackage{microtype}
\usepackage{inconsolata}
\usepackage{graphicx}
\usepackage{booktabs}
\usepackage{amsmath}
\usepackage{amssymb}
\usepackage{multirow}
\usepackage{array}
\usepackage{xcolor}
\usepackage{caption}

\title{SESSE: Sketch, Expand, Sort, Summarize, Evaluate\\
  LLM-as-Judge Evaluation via Structured Decomposition}

\author{Dae Lee, Mihai Delgeanu, Adel Youssef \\ Apple \\ \texttt{dae\_lee@apple.com}}

\begin{document}
\maketitle

%% ───────────────────────────────────────────────
%% ABSTRACT  (~130 words)
%% ───────────────────────────────────────────────
\begin{abstract}
LLM-as-judge evaluation reduces response quality assessment to a
single holistic A/B preference choice, providing no mechanism to
isolate which quality dimensions drove the preference or distinguish
model errors from genuine label ambiguity.
We propose \textbf{SESSE} (Sketch, Expand, Sort, Summarize, Evaluate),
a training-free framework that decomposes holistic judgment into
structured sub-questions mined directly from the judge's own
error cases --- requiring no oracle responses, task-specific rubrics,
or fine-tuning.
On RewardBench ($n{=}1{,}000$), SESSE achieves near-parity with the
chain-of-thought baseline and is competitive with RISE-Judge-32B (92.7\%), a
fine-tuned specialist, while remaining fully training-free.
Per-criterion vote evidence provides an interpretable audit trail for
diagnosing label ambiguity and judge failure modes unavailable from a
single holistic output token.
\end{abstract}

%% ───────────────────────────────────────────────
\section{Introduction}
%% ───────────────────────────────────────────────

LLM-as-judge evaluation has become a standard paradigm for scalable
response quality assessment --- popularized by MT-Bench
\cite{zheng2023judging}, Chatbot Arena \cite{chiang2024chatbot}, and
AlpacaFarm \cite{dubois2024alpacafarm} --- yet it reduces evaluation
to a single holistic A/B preference choice.
Even with chain-of-thought prompting, the reasoning is unstructured
free text produced in a single pass --- offering no mechanism to
decompose quality into auditable dimensions, isolate which aspects
drove the preference, or detect systematic annotation noise.

\begin{figure}[t]
  \centering
  \includegraphics[width=\columnwidth]{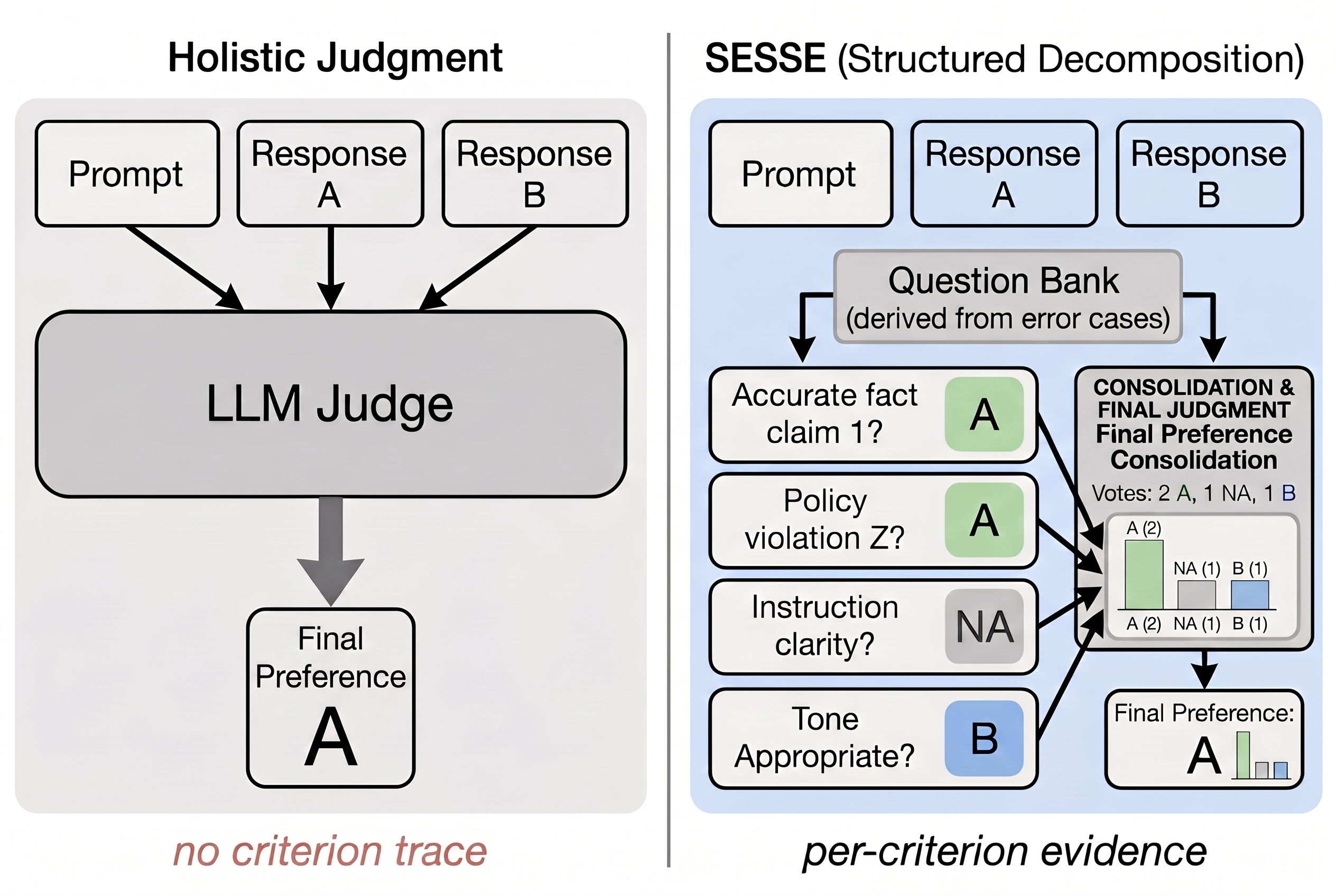}
  \caption{SESSE replaces holistic judgment with structured
    decomposition --- criteria emerge automatically from the
    judge's own error cases, producing per-criterion vote evidence.}
  \label{fig:intro}
\end{figure}

Existing approaches fall into two paradigms, each with a critical
limitation. The first, \emph{criterion-based} methods --- G-Eval
\cite{liu2023geval}, FActScore \cite{min2023factscore}, UniEval
\cite{zhong2022towards}, CheckList \cite{ribeiro2020beyond} ---
decompose quality into predefined dimensions but require
\emph{manually specified criteria} or human-designed rubrics. The
second, \emph{trained specialist} methods, fits directly to labeled
data: oracle-dependent approaches such as DeCE \cite{yu2025beyond}
and CheckEval \cite{lee-etal-2025-checkeval} derive criteria from gold
responses, while RISE-Judge \cite{risejudge2026} and other
fine-tuned variants --- JudgeLM \cite{zhu2025judgelm}, Prometheus 2
\cite{kim2024prometheus2}, Auto-J \cite{li2023generative} --- improve
calibration via SFT+DPO at the cost of training data and parameter
updates.
SESSE differs from all prior work: criteria emerge \emph{directly
from the judge's own error distribution} on standard pairwise labels,
requiring no oracle responses, task-type annotation, or fine-tuning.

We propose \textbf{SESSE}, a training-free framework that replaces
holistic judgment with structured decomposition.
Error cases from a first-pass holistic run on a dev set drive automatic
mining of evaluation criteria.
These criteria are clustered into $k$ groups and generalized into a
reusable A/B/NA (Not Applicable) sub-question bank.
Independent per-criterion voting produces an interpretable audit trail
and a final preference; ordering randomization across position
mitigates positional bias \cite{wang2023large}, and constraining the
judge to discrete tokens A/B/NA yields superior calibration over
open-ended scoring \cite{zhao2021calibrate,ren2023self}.

Decomposition addresses the reliability problem central to holistic
judgment: a single output token conflates every quality dimension into
one binary decision, so two judges --- or one judge across two runs ---
can disagree without any way to localize why. By contrast, each SESSE
sub-question is independently verifiable: an annotator, a second
model, or the same judge at a later time can re-check a single A/B/NA
vote in isolation, without re-deriving the entire holistic judgment.
This mirrors why checklist- and rubric-style protocols improve
inter-rater agreement over free-form human judgments
\cite{ribeiro2020beyond}: breaking one hard judgment into many easy
ones bounds disagreement to specific, inspectable criteria rather than
an opaque holistic vote.

Our contributions are: \textbf{(i)} SESSE, a fully automated 5-stage
pipeline that mines evaluation criteria from judge error cases,
clusters them into a reusable sub-question bank, and aggregates
per-criterion votes --- without oracle responses, rubrics, or
fine-tuning; and \textbf{(ii)} empirical evidence that SESSE is
competitive with fine-tuned specialist evaluators on RewardBench while
providing per-criterion diagnostic evidence unavailable from holistic
judgment.

%% ───────────────────────────────────────────────
\section{Method}
%% ───────────────────────────────────────────────

\begin{figure*}[t]
  \centering
  % TODO: Replace with manually designed figure (Keynote/PPT) before final submission
  \includegraphics[width=\textwidth]{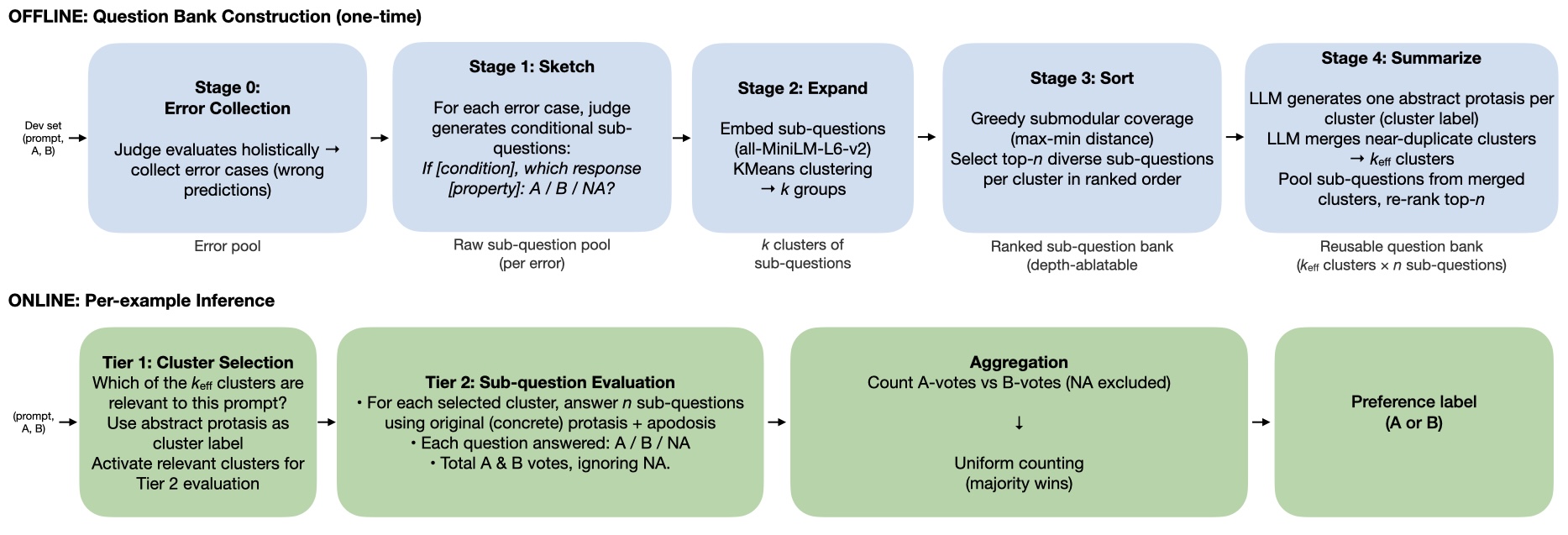}
  \caption{SESSE pipeline: offline bank construction (Stages 0--4,
    once) and online per-example inference (Stage 5).}
  \label{fig:pipeline}
\end{figure*}

SESSE operates as a five-stage pipeline (Figure~\ref{fig:pipeline}).
\textbf{Stages 0--1 (Error Collection, Sketch).}
Stage~0 runs the judge holistically on the dev set, collecting
cases where the prediction disagrees with the ground-truth label.
Stage~1 prompts the judge to generate conditional evaluation criteria
from these errors as sub-questions: \textit{``If [protasis: the
condition under which the criterion applies], which response better
satisfies [apodosis: the evaluation property]: A, B, or NA?''}
Criteria emerge from the judge's own failure distribution --- unlike
Prometheus \cite{kim2024prometheus} and G-Eval \cite{liu2023geval},
which require manually specified dimensions.

\textbf{Stages 2--3 (Expand, Sort).}
Stage~2 embeds candidate sub-questions and clusters them into $k$
thematic groups.
Stage~3 ranks sub-questions within each cluster via greedy
farthest-first traversal \cite{gonzalez1985clustering},
producing a diversity-ordered active bank of depth $n \leq n_{\max}$
without rerunning clustering, enforcing \textbf{criterion orthogonality}
across the bank.
This ranked structure enables depth ablation at inference time
(Appendix~\ref{app:ablation}).

\textbf{Stage 4 (Summarize).}
An LLM call produces one \textit{generalized} protasis per cluster,
serving as a human-readable cluster label.
Near-duplicate clusters are merged via LLM binary equivalence judgment;
sub-questions from merged clusters are pooled and
top-$n$ re-selected.

\textbf{Stage 5 (Evaluate).}
Tier~1 presents each abstract protasis to the judge and selects only
relevant clusters for the current example, skipping inapplicable
clusters to reduce per-example cost.
Tier~2 answers each selected question with its \textit{original}
(error-derived) protasis paired with its apodosis, preserving the
concrete, instance-specific phrasing from Stage~1.
When a criterion does not apply (e.g., a code cluster on a poetry
prompt), the judge returns NA; NA votes are excluded from the A/B
tally, letting the judge gracefully skip criteria that do not apply.
The final preference is the majority of all A and B ballots across
selected clusters; if A and B votes are equal, the example is declared
a tie and excluded from the accuracy denominator.

%% ───────────────────────────────────────────────
\section{Experiments}
%% ───────────────────────────────────────────────

\subsection{Setup}

\noindent\textbf{Dataset:} RewardBench \cite{lambert2024rewardbench},
2,985 pairwise preference examples across chat, chat-hard, safety,
and reasoning subsets.
Dev ($n{=}1{,}985$) for error collection; held-out val
($n{=}1{,}000$) for all reported metrics. The dev/val split is a
random partition of the full 2,985-example set: 1,000 examples held
out for val, with the remainder used as dev.

\noindent\textbf{Metric:} Pairwise preference accuracy on held-out val.
Table~\ref{tab:main} reports CoT holistic accuracy split by whether SESSE
reaches a decision: \emph{non-tie} rows (SESSE commits to a preference) and
\emph{tie} rows (equal A/B votes, SESSE has no preference to offer); Tie\% is the fraction of tied examples. SESSE accuracy is reported on
non-tie rows only.

\noindent\textbf{Models:} Qwen2-VL-7B \cite{wang2024qwen2vl}, Gemini Flash Lite, and Gemini 2.5
Flash \cite{geminiteam2024gemini15}.

\noindent\textbf{Configuration:} $k{=}25$ clusters (selected by
preliminary silhouette analysis), $n{=}10$, temperature 0.
Optimal depth $n^*$ per model is selected by dev accuracy.

\subsection{Main Results}

\begin{table}[t]
  \small
  \centering
  \begin{tabular}{@{}lcccc@{}}
    \toprule
    \textbf{System} & \multicolumn{2}{c}{\textbf{Holistic}} & \textbf{SESSE} & \textbf{Tie\%} \\
    \cmidrule(lr){2-3}
     & non-tie & tie & non-tie & \\
    \midrule
    Qwen2-VL-7B        & 0.698 & 0.667 & 0.663$^\ddagger$ & 21.3\% \\
    Flash Lite         & 0.880 & 0.679 & 0.820$^\ddagger$ &  2.6\% \\
    Gemini 2.5 Flash   & 0.945 & 0.695 & 0.933       &  3.5\% \\
    \midrule
    \shortstack[l]{RISE-Judge (32B)$^\dagger$ \\ \cite{risejudge2026}} & \multicolumn{2}{c}{0.927} & \multicolumn{2}{c}{---} \\
    \bottomrule
  \end{tabular}
  \caption{RewardBench val ($n{=}1{,}000$), dev-optimal $n^*$.
    $^\ddagger p{<}0.05$ McNemar (non-tie subset). $^\dagger$SFT+DPO.}
  \label{tab:main}
\end{table}

Table~\ref{tab:main} presents SESSE results at the dev-optimal $n^*$
across all three evaluated models. To ensure a fair comparison, all
gaps between the holistic baseline and SESSE are computed on
non-tie rows only, matching the McNemar test, which likewise
restricts its analysis to the pairs where the two methods disagree,
testing $H_0$: SESSE and the CoT holistic baseline are equally likely
to be correct when their predictions diverge.
Gemini 2.5 Flash shows a minor 1.3\% relative gap, which is not
statistically significant ($p{>}0.05$). Flash Lite and Qwen2-VL-7B
show statistically significant relative gaps of 6.8\% and 5.0\%,
respectively ($p{<}0.05$). Notably, SESSE with Gemini 2.5 Flash
reaches 93.3\% accuracy alongside a low 3.5\% tie rate, placing it in
the same performance tier as the specialist evaluator
RISE-Judge~(32B) \cite{risejudge2026} (92.7\%); while RISE-Judge
requires SFT+DPO fine-tuning, SESSE achieves comparable performance
while remaining fully training-free and providing per-criterion
evidence.

Depth ablation (Appendix~\ref{app:ablation}) shows a per-tier pattern:
the capable judge (Gemini 2.5 Flash) saturates at $n{=}2$
sub-questions per cluster, the mid-tier judge (Flash Lite) benefits
from broader evidence and peaks at $n{=}5$, while the weakest judge
(Qwen2-VL-7B) peaks earlier, at $n{=}3$, before degrading.
Qwen2-VL-7B's degradation past $n{=}3$ reveals a bank quality effect: accuracy
peaks at $n{=}3$ then degrades, because 51.9\% of instances are covered
by a single dominant protasis in the self-generated bank --- beyond
$n{=}3$, the depth budget is consumed by redundant sub-questions from
that cluster.
A cross-model transfer experiment --- regenerating the bank with Gemini
while keeping Qwen2-VL-7B as judge --- confirms the accuracy ceiling is
set by judge capability, not bank quality
(Table~\ref{tab:config_full}, Appendix~\ref{app:config}).

SESSE tie rows are harder examples: holistic accuracy on tie rows
is consistently lower than on non-tie rows across all three models
(Table~\ref{tab:main}).
The gap is most pronounced for capable judges --- Gemini 2.5 Flash
drops from 94.5\% to 69.5\% on tie rows, suggesting that ties
signal genuine ambiguity rather than arbitrary disagreement.

%% ───────────────────────────────────────────────
\section{Discussion}
\label{sec:discussion}
%% ───────────────────────────────────────────────

\textbf{Training-free parity with fine-tuned judges.}
SESSE's near-parity with CoT holistic for Gemini 2.5 Flash
(Table~\ref{tab:main}) represents a minor trade-off: near-parity in
exchange for structured, auditable per-criterion evidence at no
training cost. CoT holistic itself already outperforms
RISE-Judge \cite{risejudge2026}, so SESSE's comparable accuracy
mainly shows that it preserves the base model's advantage without
requiring fine-tuning.
For mid-tier judges, the gap is significant but reflects a different
dynamic: Flash Lite's drop stems primarily from Tier-1 false positives
on safety subsets (see \textbf{Domain-specific signal recovery} below),
not from the decomposition mechanism itself.

\textbf{Domain-specific signal recovery.}
SESSE's impact varies substantially across RewardBench subsets.
On \texttt{donotanswer}, Gemini 2.5 Flash gains +20.6\% relative
over holistic ($75.9\% \rightarrow 91.5\%$) as explicit
safety-refusal criteria isolate a signal that holistic judgment
conflates with general preference.
Conversely, \texttt{xstest-should-respond} drops 27.1\% relative due
to Tier-1 false positives on borderline-safe prompts, pointing to
Tier-1 precision as the key failure mode for domain-sensitive subsets.

\textbf{When to prefer SESSE.}
SESSE is preferable when interpretability matters: the per-criterion
vote trail enables annotation auditing and exposes label noise,
justifying its inference overhead beyond raw accuracy.

\textbf{Tie rate as a deployment-time capability signal.}
Tie\% is a label-free proxy for judge reliability: Qwen2-VL-7B's
21.3\% tie rate --- versus 2.6--3.5\% for the Gemini models ---
predicts its accuracy gap, since the judge cannot form a directional
preference on roughly one-in-five examples (which score only 66.7\%
accuracy when forced holistically; Table~\ref{tab:main}).
A high Tie\% on a held-out probe thus signals a capability mismatch or
domain shift without requiring ground-truth labels, letting a
practitioner switch judges, retarget the bank, or flag examples for
human review.

\textbf{Failure modes.}
When SESSE diverges from holistic judgment, analysis reveals
capability-dependent failure modes (see Table~\ref{tab:degradation},
Appendix~\ref{app:degradation}, for the full three-bucket taxonomy).
We conducted further analysis on Flash Lite and Gemini 2.5 Flash
(SESSE accuracy above 80\%) to understand failure modes.
For Flash Lite, disagreements split roughly evenly between genuine
label ambiguity (49\%, high vote entropy), decomposition failure
(26\%), and semantic equivalence (26\%), reflecting a mid-tier judge
without a single dominant failure signature.
For the capable judge (Gemini 2.5 Flash), 48\% of disagreements instead
reflect \emph{semantic equivalence} between responses (C3): these
pairs have higher response similarity than the dataset average
(0.75 vs.\ 0.69 mean cosine similarity between Sentence-BERT
embeddings; \citealt{reimers2019sentencebert}),
and the bank correctly returns high NA rates when no meaningful
distinction exists.
This C3 signature is not a model failure but a dataset property ---
nearly half of Gemini's degradation rows encode stylistic preference
rather than substantive quality differences, making the per-criterion
vote distribution a lightweight benchmark audit tool independent of
accuracy measurement.
In all cases, per-criterion vote distributions make failure modes
explicitly diagnosable, a feature absent from monolithic A/B scoring.

%% ───────────────────────────────────────────────
\section{Conclusion}
%% ───────────────────────────────────────────────

We presented SESSE, a training-free framework that decomposes holistic
LLM-as-judge evaluation into automatically derived A/B/NA sub-questions
mined from the judge's own error distribution. On RewardBench, SESSE
achieves a non-significant gap to CoT holistic for capable judges and
is in the same performance range as a fine-tuned specialist evaluator
(RISE-Judge 32B) while remaining training-free. Beyond accuracy, the
per-criterion vote distributions provide interpretable diagnostic
evidence --- surfacing failure modes and benchmark quality issues that
holistic scoring leaves opaque --- a use case that scales with judge
capability rather than against it.

%% ───────────────────────────────────────────────
\section*{Limitations}
%% ───────────────────────────────────────────────

\textbf{Bank portability.}
The question bank is derived from a specific judge model's error
distribution on a specific dev set.
Changing the judge model or evaluation domain requires re-running
Stages 0--4 (a one-time offline cost).
Amortization strategies --- cross-domain bank transfer, incremental
error pool updates, shared banks across related judge models --- are
not explored here.

\textbf{Cluster count \& benchmark scope.}
We fix $k{=}25$ by silhouette analysis; ablation over $k$ and
generalization beyond RewardBench (open-ended generation, multi-turn,
non-English) are deferred to future work.

\textbf{Inference cost.}
SESSE at $n{=}10$ requires ${\sim}282$ LLM calls per example
vs.\ 1 holistic; capable judges saturate at $n{=}2$, reducing cost
proportionally with no accuracy loss.

\textbf{Decomposition limits.}
A/B/NA sub-questions cannot capture end-to-end execution evaluation
(code compilation, derivation correctness) --- the primary source of
SESSE degradation on reasoning and code subsets.

%% ───────────────────────────────────────────────
\section*{Ethical Considerations}
%% ───────────────────────────────────────────────

SESSE evaluates LLM outputs using an LLM judge, inheriting any
biases present in the judge model's error distribution.
Criteria mined from judge failures may encode systematic biases ---
users deploying SESSE in high-stakes annotation pipelines should
audit the generated question bank for bias before use.
All experiments use publicly available models and datasets.

\bibliography{references_emnlp}

@article{wang2024qwen2vl,
  title={Qwen2-{VL}: Enhancing Vision-Language Model's Perception of the World at Any Resolution},
  author={Wang, Peng and others},
  journal={arXiv preprint arXiv:2409.12191},
  year={2024}
}

@article{geminiteam2024gemini15,
  title={Gemini 1.5: Unlocking Multimodal Understanding Across Millions of Tokens of Context},
  author={{Gemini Team}},
  journal={arXiv preprint arXiv:2403.05530},
  year={2024}
}

@article{gonzalez1985clustering,
  title={Clustering to minimize the maximum intercluster distance},
  author={Gonzalez, Teofilo F.},
  journal={Theoretical Computer Science},
  volume={38},
  pages={293--306},
  year={1985},
  publisher={Elsevier}
}

@inproceedings{zheng2023judging,
  title={Judging {LLM}-as-a-Judge with {MT-Bench} and Chatbot Arena},
  author={Zheng, Lianmin and Chiang, Wei-Lin and Sheng, Ying and Zhuang, Siyuan and Wu, Zhanghao and Zhuang, Yonghao and Lin, Zi and Li, Zhuohan and Li, Dacheng and Xing, Eric and others},
  booktitle={Advances in Neural Information Processing Systems},
  year={2023}
}

@inproceedings{dubois2024alpacafarm,
  title={{AlpacaFarm}: A Simulation Framework for Methods that Learn from Human Feedback},
  author={Dubois, Yann and Li, Xuechen and Taori, Rohan and Zhang, Tianyi and Gulrajani, Ishaan and Ba, Jimmy and Guestrin, Carlos and Liang, Percy and Hashimoto, Tatsunori B},
  booktitle={Advances in Neural Information Processing Systems},
  year={2023}
}

@inproceedings{kim2024prometheus2,
  title={Prometheus 2: An Open Source Language Model Specialized in Evaluating Other Language Models},
  author={Kim, Seungone and Suk, Juyoung and Longpre, Shayne and Lin, Bill Yuchen and Shin, Jamin and Welleck, Sean and Neubig, Graham and Lee, Moontae and Lee, Kyungjae and Seo, Minjoon},
  booktitle={Proceedings of the 2024 Conference on Empirical Methods in Natural Language Processing},
  year={2024}
}

@article{lambert2024rewardbench,
  title={{RewardBench}: Evaluating Reward Models for Language Modeling},
  author={Lambert, Nathan and Pyatkin, Valentina and Morrison, Jacob and Miranda, LJ and Lin, Bill Yuchen and Chandu, Khyathi and Dziri, Nouha and Kumar, Sachin and Zick, Tom and Choi, Yejin and others},
  journal={arXiv preprint arXiv:2403.13787},
  year={2024}
}

@inproceedings{li2023generative,
  title={Generative Judge for Evaluating Alignment},
  author={Li, Junlong and Sun, Shichao and Yuan, Weizhe and Fan, Run-Ze and Zhao, Hai and Liu, Pengfei},
  booktitle={Proceedings of the 12th International Conference on Learning Representations},
  year={2024}
}

@inproceedings{liu2023geval,
  title={{G-Eval}: {NLG} Evaluation using GPT-4 with Better Human Alignment},
  author={Liu, Yang and Iter, Dan and Xu, Yichong and Wang, Shuohang and Xu, Ruochen and Zhu, Chenguang},
  booktitle={Proceedings of the 2023 Conference on Empirical Methods in Natural Language Processing},
  year={2023}
}

@inproceedings{min2023factscore,
  title={{FActScore}: Fine-grained Atomic Evaluation of Factual Precision in Long Form Text Generation},
  author={Min, Sewon and Krishna, Kalpesh and Lyu, Xinxi and Lewis, Mike and Yih, Wen-tau and Koh, Pang Wei and Iyyer, Mohit and Zettlemoyer, Luke and Hajishirzi, Hannaneh},
  booktitle={Proceedings of the 2023 Conference on Empirical Methods in Natural Language Processing},
  year={2023}
}

@inproceedings{ren2023self,
  title={Self-Evaluation Improves Selective Generation in Large Language Models},
  author={Ren, Jie and Zhao, Yao and Vu, Tu and Liu, Peter J and Lakshminarayanan, Balaji},
  booktitle={NeurIPS 2023 Workshop on Robustness of Few-shot and Zero-shot Learning in Foundation Models},
  year={2023}
}

@inproceedings{ribeiro2020beyond,
  title={Beyond Accuracy: Behavioral Testing of {NLP} Models with {CheckList}},
  author={Ribeiro, Marco Tulio and Wu, Tongshuang and Guestrin, Carlos and Singh, Sameer},
  booktitle={Proceedings of the 58th Annual Meeting of the Association for Computational Linguistics},
  year={2020}
}

@inproceedings{risejudge2026,
  title={Improve {LLM}-as-a-Judge Ability as a General Ability},
  author={Yu, Jiachen and Sun, Shaoning and Hu, Xiaohui and Yan, Jiaxu and Yu, Kaidong and Li, Xuelong},
  booktitle={Proceedings of the 2025 Conference on Empirical Methods in Natural Language Processing},
  month={nov},
  year={2025},
  address={Suzhou, China},
  publisher={Association for Computational Linguistics},
  url={https://aclanthology.org/2025.emnlp-main.712/},
  doi={10.18653/v1/2025.emnlp-main.712},
  pages={14099--14115}
}

@inproceedings{wang2023large,
  title={Large Language Models are not Robust Multiple Choice Selectors},
  author={Wang, Peiyi and Li, Lei and Chen, Liang and Zhu, Dawei and Lin, Binghuai and Cao, Yunbo and Liu, Qi and Liu, Tianyu and Sui, Zhifang},
  booktitle={Proceedings of the 12th International Conference on Learning Representations},
  year={2024}
}

@inproceedings{yu2025beyond,
  title={Beyond Accuracy: Ensuring Correct Format and Content in Mathematical Reasoning via Dual Decomposition},
  author={Yu, Jiachen and others},
  booktitle={Proceedings of the 2025 Conference on Empirical Methods in Natural Language Processing},
  year={2025}
}

@inproceedings{zhao2021calibrate,
  title={Calibrate Before Use: Improving Few-shot Performance of Language Models},
  author={Zhao, Zihao and Wallace, Eric and Feng, Shi and Klein, Dan and Singh, Sameer},
  booktitle={Proceedings of the 38th International Conference on Machine Learning},
  year={2021}
}

@inproceedings{zhong2022towards,
  title={Towards a Unified Multi-Dimensional Evaluator for Text Generation},
  author={Zhong, Ming and Liu, Yang and Chen, Da and Peng, Danqing and Xiao, Pengfei and Wang, Dongyu and Chen, Jiawei and Gu, Jialong and Liang, Yilun and others},
  booktitle={Proceedings of the 2022 Conference on Empirical Methods in Natural Language Processing},
  year={2022}
}

@inproceedings{zhu2025judgelm,
  title={{JudgeLM}: Fine-tuned Large Language Models are Scalable Judges},
  author={Zhu, Lianghui and Wang, Xinggang and Wang, Xinlong},
  booktitle={Proceedings of the 38th AAAI Conference on Artificial Intelligence},
  year={2024}
}

@inproceedings{chiang2024chatbot,
  title={Chatbot Arena: An Open Platform for Evaluating {LLMs} by Human Preference},
  author={Chiang, Wei-Lin and Zheng, Lianmin and Sheng, Ying and Angelopoulos, Anastasios Nikolas and Li, Tianle and Li, Dacheng and Zhang, Hao and Zhu, Banghua and Jordan, Michael and Gonzalez, Joseph E and others},
  booktitle={Proceedings of the 41st International Conference on Machine Learning},
  year={2024}
}

@inproceedings{kim2024prometheus,
  title={Prometheus: Inducing Fine-grained Evaluation Capability in Language Models},
  author={Kim, Seungone and Shin, Jamin and Cho, Yejin and Jang, Joel and Longpre, Shayne and Lee, Hwaran and Yun, Sangdoo and Shin, Seongjin and Kim, Sungdong and Thorne, James and Seo, Minjoon},
  booktitle={Proceedings of the Twelfth International Conference on Learning Representations},
  year={2024}
}

@inproceedings{lee-etal-2025-checkeval,
    title = "{C}heck{E}val: A reliable {LLM}-as-a-Judge framework for evaluating text generation using checklists",
    author = "Lee, Yukyung  and
      Kim, JoongHoon  and
      Kim, Jaehee  and
      Cho, Hyowon  and
      Kang, Jaewook  and
      Kang, Pilsung  and
      Kim, Najoung",
    editor = "Christodoulopoulos, Christos  and
      Chakraborty, Tanmoy  and
      Rose, Carolyn  and
      Peng, Violet",
    booktitle = "Proceedings of the 2025 Conference on Empirical Methods in Natural Language Processing",
    month = nov,
    year = "2025",
    address = "Suzhou, China",
    publisher = "Association for Computational Linguistics",
    url = "https://aclanthology.org/2025.emnlp-main.796/",
    doi = "10.18653/v1/2025.emnlp-main.796",
    pages = "15771--15798",
    ISBN = "979-8-89176-332-6"
}

@inproceedings{reimers2019sentencebert,
  title={Sentence-{BERT}: Sentence Embeddings using Siamese {BERT}-Networks},
  author={Reimers, Nils and Gurevych, Iryna},
  booktitle={Proceedings of the 2019 Conference on Empirical Methods in Natural Language Processing},
  year={2019}
}

%% ───────────────────────────────────────────────
\appendix
\renewcommand{\thetable}{\Alph{section}.\arabic{table}}
\setcounter{table}{0}
%% ───────────────────────────────────────────────

\section{Depth Ablation}
\label{app:ablation}

Table~\ref{tab:ablation} sweeps $n{=}1$--$10$ sub-questions per cluster
for all three self-generated banks (bold = best per model).
All models plateau quickly: accuracy stabilizes within the first few
values of $n$ and shows little change beyond that, suggesting a small
question budget is sufficient once the most informative sub-questions
are exhausted.
Qwen2-VL-7B is the exception, showing a slight degradation at larger
$n$ due to limited cluster diversity in the self-generated bank.

\begin{table*}[!ht]
  \small
  \centering
  \begin{tabular}{lccc}
    \toprule
    \textbf{$n$} & \textbf{Qwen2-VL-7B} & \textbf{Gemini 2.5 Flash} & \textbf{Flash Lite} \\
    \midrule
    1  & 0.626 & 0.923 & 0.790 \\
    2  & 0.648 & \textbf{0.933} & 0.804 \\
    3  & \textbf{0.663} & 0.927 & 0.801 \\
    5  & 0.644 & 0.932 & \textbf{0.820} \\
    7  & 0.645 & 0.929 & 0.817 \\
    10 & 0.643 & 0.930 & 0.819 \\
    \bottomrule
  \end{tabular}
  \caption{Accuracy (excl.\ ties) vs.\ $n$ (self-generated banks; bold = $n^*$).}
  \label{tab:ablation}
\end{table*}

\setcounter{table}{0}
\section{Cross-Model Bank Transfer}
\label{app:config}

Table~\ref{tab:config_full} compares Qwen2-VL-7B accuracy when evaluated
with a self-generated bank versus a bank whose questions were generated by
Gemini 2.5 Flash, holding the judge model fixed.
Both banks reach the same peak accuracy (66.3\%), showing that
substituting a stronger model's questions does not meaningfully
change evaluation performance when the judge model is held fixed.

\begin{table*}[!t]
  \small
  \centering
  \begin{tabular}{lccc}
    \toprule
    \textbf{Config} & $k_\text{eff}$ & \textbf{SESSE acc ($n{=}10$)} & \textbf{SESSE ($n^*$)} \\
    \midrule
    Qwen self-generated &  8 & 0.640 & 0.663 ($n^*{=}3$) \\
    Qwen$+$Gemini bank  & 14 & 0.637 & 0.663 ($n^*{=}2$) \\
    \bottomrule
  \end{tabular}
  \caption{Qwen2-VL-7B with Gemini-generated vs.\ self-generated bank (val).}
  \label{tab:config_full}
\end{table*}

\setcounter{table}{0}
\section{Failure Mode Taxonomy}
\label{app:degradation}

Table~\ref{tab:degradation} reports the per-configuration breakdown of
degradation rows --- examples where SESSE and the holistic baseline disagree.
We report this breakdown only for configurations with SESSE accuracy
above 80\% (Flash Lite, Gemini 2.5 Flash); at lower accuracy, degradation
rows are too frequent to characterize as a small set of distinct
failure signatures.
Each row is attributed to one of three buckets:
\textbf{C1 (Ambiguous)} reflects genuine label difficulty, where sub-question
votes are evenly split;
\textbf{C2 (Decomposition failure)} indicates the judge is confidently wrong,
with $\geq$65\% of non-NA votes against the ground-truth label;
\textbf{C3 (Semantic equivalence)} indicates the bank correctly declines to pick a winner,
returning high NA rates when the two responses are too similar to discriminate.

\begin{table*}[!t]
  \small
  \centering
  \resizebox{\textwidth}{!}{%
  \begin{tabular}{lcccccc}
    \toprule
    \textbf{Config} & \textbf{Judge} & \textbf{Bank} & \textbf{Degradation (\% val)} & \textbf{C1 (Ambiguous)} & \textbf{C2 (Decomposition)} & \textbf{C3 (Semantic equiv.)} \\
    \midrule
    Flash Lite         & Flash Lite        & self-generated   & 105 (10.5\%) & 49\% & 26\%          & 26\% \\
    Gemini 2.5 Flash   & Gemini 2.5 Flash  & self-generated   &  48  (4.8\%) & 29\% & 23\%          & \textbf{48\%} \\
    \bottomrule
  \end{tabular}%
  }
  \caption{Degradation bucket distribution. C1 = genuine ambiguity
    (high vote entropy); C2 = decomposition failure (SESSE
    consistently wrong $\geq$65\% non-NA votes against label); C3 = semantic
    equivalence (high NA rate, responses too similar to discriminate).
    Percentages of degradation rows per config.}
  \label{tab:degradation}
\end{table*}

\end{document}